%% file: main.tex
\input{_constants}
 \arxiv

\pdfoutput=1
\documentclass[10pt,twocolumn,letterpaper]{article}
\input{cvpr_header}

\unless\ifarxiv \myexternaldocument{_supplementary} \fi

\begin{document}
%% TITLE
\title{Learned Image Compression with Dictionary-based Entropy Model}
\author{\authorBlock}
\maketitle

\input{00_abstract}
\input{01_intro}
\input{02_related}

\input{03_method}

\input{04_experiments}
\input{10_conclusion}
\input{05_acknowledgment}

{\small
\bibliographystyle{ieeenat_fullname}
\bibliography{11_references}
}

\ifarxiv \clearpage \appendix \input{12_appendix} \fi

\end{document}

%% file: _constants.tex
\def\paperID{1172}
\def\confName{CVPR}
\def\confYear{2025}

\def\authorBlock{
    Jingbo Lu\textsuperscript{1} \qquad
    Leheng Zhang\textsuperscript{1}\qquad
    Xingyu Zhou\textsuperscript{1} \qquad
    Mu Li\textsuperscript{2} \qquad
    Wen Li\textsuperscript{1} \qquad
    Shuhang Gu\textsuperscript{1}\thanks{Corresponding author.}\\
    \textsuperscript{1} University of Electronic Science and Technology of China \\
    \textsuperscript{2} Harbin Institute of Technology, Shenzhen
    \\
    {\tt\small \{jingbolu2023, shuhanggu\}@gmail.com}
    \\
    {\tt\small \href{https://github.com/LabShuHangGU/DCAE}{https://github.com/LabShuHangGU/DCAE}}
    
}

\newif\ifreview 
\newif\ifarxiv \newcommand{\arxiv}{\arxivtrue}
\newif\ifcamera 
\newif\ifrebuttal 

%% file: cvpr_header.tex
\ifreview \usepackage[review]{cvpr} \fi
\ifarxiv \usepackage[pagenumbers]{cvpr} \fi
\ifrebuttal \usepackage[rebuttal]{cvpr} \fi
\ifcamera \usepackage{cvpr} \fi

\input{_macros}  % Add packages to _macros.tex

\usepackage{xr-hyper}

\makeatletter
\newcommand*{\addFileDependency}[1]{
  \typeout{(#1)}
  \@addtofilelist{#1}
  \IfFileExists{#1}{}{\typeout{No file #1.}}
}

\makeatother
\newcommand*{\myexternaldocument}[1]{
    \externaldocument{#1}
    \addFileDependency{#1.tex}
    \addFileDependency{#1.aux}
}

\definecolor{cvprblue}{rgb}{0.21,0.49,0.74}
\usepackage[pagebackref,breaklinks,colorlinks,citecolor=cvprblue]{hyperref}
\usepackage[capitalize]{cleveref}
\crefname{section}{Sec.}{Secs.}
\crefname{table}{Table}{Tables}
\crefname{figure}{Fig.}{Figs.}

\ifarxiv \crefname{appendix}{App.}{Apps.}
\else \crefname{appendix}{Suppl.}{Suppls.} \fi

%% file: _macros.tex
\usepackage{graphicx}	
\usepackage{amsmath}	
\usepackage{amssymb}	
\usepackage{booktabs}
\usepackage{times}
\usepackage{microtype}
\usepackage{epsfig}
\usepackage[table,xcdraw,dvipsnames]{xcolor}
\usepackage{caption}
\usepackage{float}
\usepackage{placeins}
\usepackage{color, colortbl}
\usepackage{stfloats}
\usepackage{enumitem}
\usepackage{tabularx}
\usepackage{xstring}
\usepackage{multirow}
\usepackage{xspace}
\usepackage{url}
\usepackage{subcaption}
\usepackage{xcolor}
\usepackage{amsmath}
\usepackage{bm}
\usepackage[accsupp]{axessibility}

\usepackage[hang,flushmargin]{footmisc}

\ifcamera \usepackage[accsupp]{axessibility} \fi

% ...

\ifarxiv  \fi

\newcommand{\R}[1]{{%
    \textbf{%
        \ifstrequal{#1}{1}{\textcolor{red}{R#1}}{%
        \ifstrequal{#1}{2}{\textcolor{blue}{R#1}}{%
        \ifstrequal{#1}{3}{\textcolor{magenta}{R#1}}{%
        \ifstrequal{#1}{4}{\textcolor{teal}{R#1}}{%
                           \textcolor{cyan}{R#1}%
        }}}}%
    }%
}}

%% file: 00_abstract.tex
\begin{abstract}
% Abstract goes here.
Learned image compression methods have attracted great research interest and exhibited superior rate-distortion performance to the best classical image compression standards of the present.
The entropy model plays a key role in learned image compression, which estimates the probability distribution of the latent representation for further entropy coding.
Most existing methods employed hyper-prior and auto-regressive architectures to form their entropy models.
However, they only aimed to explore the internal dependencies of latent representation while neglecting the importance of extracting prior from training data.
In this work, we propose a novel entropy model named Dictionary-based Cross Attention Entropy model, which introduces a learnable dictionary to summarize the typical structures occurring in the training dataset to enhance the entropy model.
Extensive experimental results have demonstrated that the proposed model strikes a better balance between performance and latency, achieving state-of-the-art results on various benchmark datasets.
\end{abstract}

%% file: 01_intro.tex
\section{Introduction}
\label{sec:intro}
Image compression is a vital and well-established research area in the field of image signal processing.
The substantial demand for high-resolution images requires powerful compression techniques to address storage and transmission challenges.
Classical standards such as JPEG \cite{wallace1991jpeg}, JPEG2000 \cite{Taubman_2002}, and VVC \cite{Rao_Dominguez_2022} have been widely adopted over time, following a general pipeline: transforming, quantization, and entropy coding. 
Recently, the learned image compression (LIC) methods \cite{he2022elic, li2023frequency, liu2023learned, wang2022neural, zhu2021transformer, zou2022devil, mentzer2023m2t, jiang2023mlic, jiang2023mlicpp, fu2024weconvene, han2024causal} have demonstrated outstanding performance, even surpassing the current best image and video coding standards VVC.

Learned image compression primarily consists of two key components: nonlinear auto-encoder and entropy model. 
\begin{figure}[]
    \centering
    \includegraphics[width=0.48\textwidth]{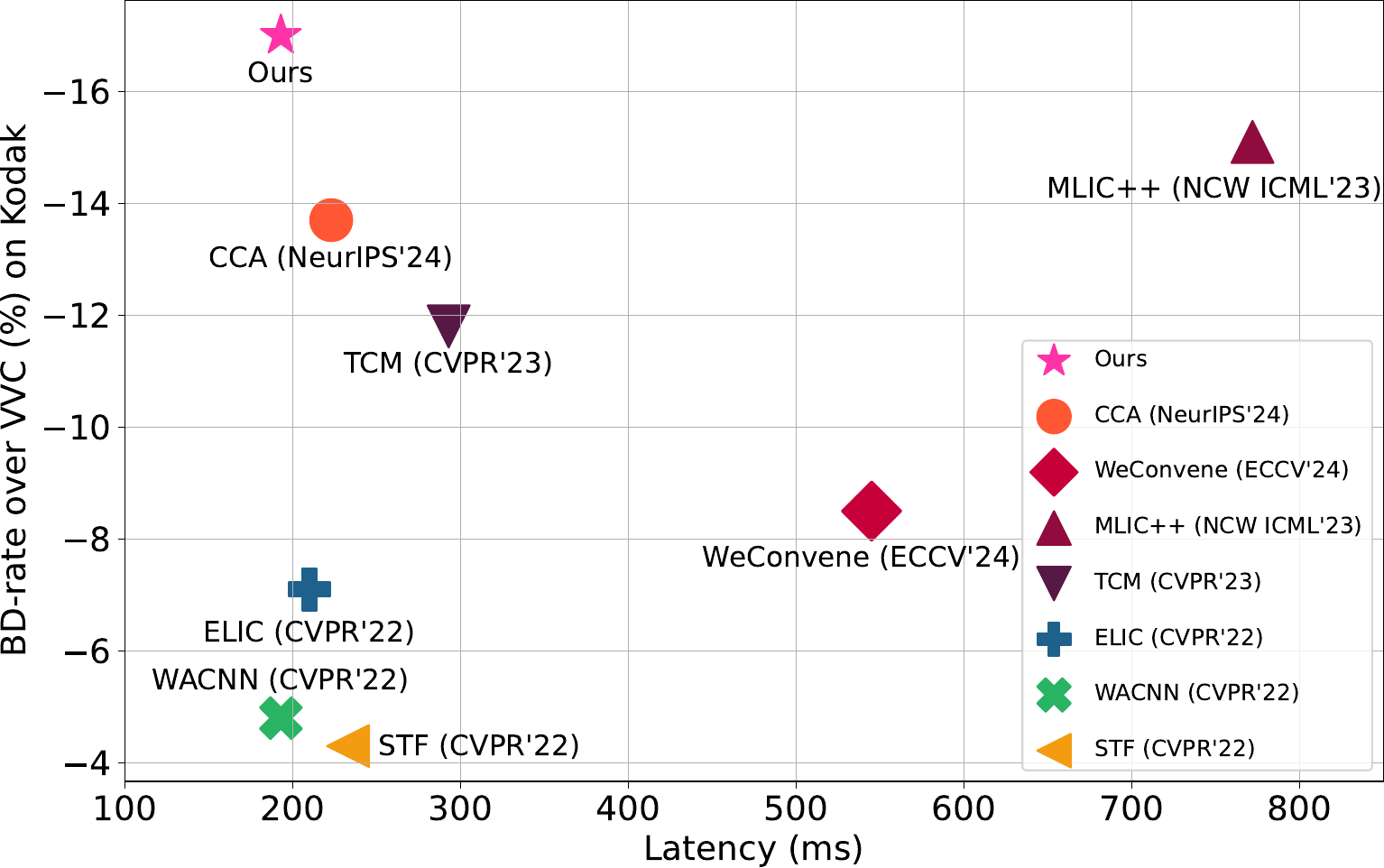}
    \caption{ Rate-speed comparison on Kodak. Left-top is better.}
    \label{fig:latency_bd_rate}
    \vspace{-10pt}
\end{figure}
The encoder and decoder play the role of transforming the image and the latent representation into each other;
while, the entropy model estimates the probability distribution of the latent  for entropy coding \cite{martin1979range, Marpe_Schwarz_Wiegand_2003}.
In the last several years, for the pursuit of better performance, one category of studies \cite{liu2023learned, li2023frequency, zou2022devil, zhu2021transformer, xie2021enhanced, cheng2020learned} investigates advanced auto-encoder structures to establish delicate latent representation extractors.
While, another line of research \cite{balle2018variational, minnen2018joint, cheng2020learned, he2021checkerboard, minnen2020channel, qian2020learning, kim2022joint, qian2022entroformer, koyuncu2022contextformer, mentzer2023m2t, han2024causal} pays attention to the more fundamental and unique component of LIC, i.e. the entropy model, to optimize the rate and distortion (RD) trade-off.

At the core of the entropy model in LIC framework is a distribution estimator.
In their seminal work, Ballé \textit{et al.}~\cite{balle2016end} showed that 
the smallest code length of latent representation is given by the cross entropy between a parameterized distribution predictor (entropy model) and the real distribution of latents, and used a  fully factorized density model to capture the probability distributions of the latent representation.
%Afterwards, 
Subsequently,
Ballé \textit{et al.}~\cite{balle2018variational} and Minnen \textit{et al.}~\cite{minnen2018joint} respectively introduced the hyper-prior and the auto-regressive frameworks, which leverage side-information or decoded representation to provide image dependent priors for better capturing the distribution of latent representation.
The great success achieved by the hyperprior and the autoregressive framework has inspired numerous follow-up works, introducing priors to establish a conditional entropy model has become a prevailing strategy in the literature of LIC.
In the last several years, numerous attempts  have been made in establishing elaborate causal context models \cite{ cheng2020learned, he2021checkerboard, minnen2020channel, qian2020learning, kim2022joint, qian2022entroformer, koyuncu2022contextformer, mentzer2023m2t} as well as designing sophisticated network architectures \cite{liu2023learned, li2023frequency, zou2022devil, zhu2021transformer, xie2021enhanced, cheng2020learned} 
for better exploiting internal dependency among latent representation.

In this paper, instead of further exploring advanced internal dependency modeling architectures, we propose a Dictionary-based Cross Attention Entropy model (DCAE) which tries to take additional advantages from external training data to improve the entropy model in the LIC framework.
%To be more specific, 
In the existing LIC works, training data helps the entropy model indirectly by providing examples to learn functions for capturing internal dependency,
little attention has been paid to directly extracting priors from training data to boost the performance of entropy model.
Nevertheless, the training data contains prior of natural images, which has proven to be effective in providing supplementary information to recover the image from corrupted observations in the field of image restoration.
Therefore, exploiting external prior for boosting entropy modeling could be a promising direction.
In order to leverage typical patterns in natural image, we learn a dictionary from the training dataset to summarize typical structures.
Then, during the auto-regressive predicting process, decoded representation, which contain partial information of local structure, can work as query tokens to select similar dictionary entries to predict the remaining representation in a cross-attention manner.
In addition, we utilize feature maps from different convolutional layers to achieve multi-scale texture extraction for helping to achieve more accurate dictionary queries.
% These innovations enable our model to outperform existing state-of-the-art methods. 
In summary, our contributions can be summarized as follows:
\begin{itemize}
    \item We apply a dictionary to summarize useful information from the training dataset. During the auto-regressive predicting process, we can use the decoded representation, which contains partial information of local structure, to select similar dictionary entries to assist in predicting  the remaining representation in a cross-attention manner.
    \item We leverage features with different receptive fields to capture textures at various scales, allowing the model to capture fine-grained and coarse-level texture information, which helps us perform more precise dictionary queries.
    \item Experiments show that our method achieves state-of-the-art performance regarding coding performance and running speed(Fig.\ref{fig:latency_bd_rate} and Tab. \ref{tab:compare}). 
\end{itemize}

%% file: 02_related.tex
\section{Related Work}
\label{sec:related}
\subsection{Learned Image Compression}
\paragraph{Learned nonlinear transforms.}
The powerful nonlinear transformation capability is one of the keys to the learned image compression. 
It functions as an encoder or decoder, converting input images into compact latent representation for entropy coding, or transforming latent representation back into reconstructed images.
The development of learned nonlinear transforms can be categorized into two types: CNN-based methods and transformer-based methods.
Since Ballé \textit{et al.} \cite{Ballé_Laparra_Simoncelli_2015} first proposed the  GDN to reduce the mutual information between features,
the combination of GDN and CNN has been widely used in subsequent methods \cite{Ballé_Laparra_Simoncelli_2016, balle2018variational, minnen2018joint}.
In order to further enhance the nonlinear transformation capability of the CNN, 
Cheng \textit{et al.} \cite{cheng2020learned} adopted the attention module to deal with the challenging content,
whereas Xie \textit{et al.} \cite{xie2021enhanced} proposed the enhanced invertible encoding network to mitigate the problem of information loss.
Recently, numerous efforts have been made to explore the transformer architecture \cite{vaswani2017attention}. 
% due to its efficacy in establishing long-range dependencies.
Zou \textit{et al.} \cite{zou2022devil} and Zhu \textit{et al.} \cite{zhu2021transformer} first employed the Swin transformer layer \cite{liu2021swin} to construct their encoder and decoder.
Zou \textit{et al.} \cite{zou2022devil} further proposed a window attention module to enhance their CNN-based model by focusing on spatially neighboring elements.
Liu \textit{et al.} \cite{liu2023learned} incorporated CNN and transformer as a fundamental module.
Furthermore, Li \textit{et al.} \cite{li2023frequency} utilized different window sizes of the Swin transformer layer to capture various frequency information.
Inspired by \cite{he2016deep} and \cite{liang2021swinir}, we use residual blocks and residual Swin transformer blocks to construct our encoder and decoder for capturing local and non-local information simultaneously.
Additionally, we further improve our transformer by using advanced techniques such as ResScale \cite{yu2023metaformer} and  Convolutional Gated Linear Unit \cite{shi2024transnext}.
\paragraph{Entropy Model.}
Another crucial module in learned image compression is the entropy model, which is used to evaluate the distribution parameters of latent representation for entropy coding.
Existing entropy models primarily pay attention to the investigation of how to capture the internal dependencies among latent representation.
Ballé \textit{et al.} \cite{balle2018variational} first proposed the hyper-prior to capture spatial internal dependencies of the latent representation.
Minnen \textit{et al.} \cite{minnen2018joint} then proposed the serial auto-regressive context model that utilizes the adjacent decoded latent representation to assist the distribution estimating of current latent representation.
In order to capture long-range dependencies of latent representation,
Qian \textit{et al.} \cite{qian2020learning} used the most relevant latent representation and
Kim \textit{et al.} \cite{kim2022joint} utilized the attention mechanism of transformer to capture global information.
Furthermore, to address the issue of slow encoding and decoding in serial auto-regressive context models, some studies \cite{he2021checkerboard, minnen2020channel, he2022elic, mentzer2023m2t} set the internal dependencies of the auto-regressive context model to a fixed order to achieve parallelization.
In addition, different sophisticated network architectures \cite{koyuncu2022contextformer, qian2022entroformer, liu2023learned, li2023frequency} were also designed to enhance the ability to model internal dependencies.
Despite various advancements aimed at improving the modeling capabilities of entropy model, they mainly focused on utilizing the internal dependencies of latent representation and neglected the exploration of prior information in  external training data.
Different from these works, the proposed Dictionary-based Cross Attention Entropy model concentrates on capturing typical patterns and textures from training dataset to establish external dependencies between the latent representation and prior derived from training dataset.
Therefore, our entropy model can achieve more accurate distribution estimation for entropy coding.
\subsection{Dictionary Learning}
Dictionary learning has demonstrated powerful potential in the fields of image generation and image restoration due to its ability of effectively utilizing prior information from the training dataset.
In image generation tasks, Van \textit{et al.} \cite{van2017neural} first introduced a dictionary to generate clear images.
In order to achieve visually pleasing generation quality, Esser \textit{et al.} \cite{esser2021taming} employed perceptual and adversarial losses to train the dictionary.
In image restoration tasks, Gu \textit{et al.}~\cite{gu2022vqfr} used a dictionary for face restoration.
Liu \textit{et al.}~\cite{liu2023learning}  further learned a set of basis dictionaries from different types of datasets for obtaining  more flexible and expressive prior.
In addition, Zhang \textit{et al.}~\cite{zhang2024transcending} employed a dictionary to study various cluster centers, enabling self-attention operations on tokens of the same category.
In learned image compression, Minnen \textit{et al.}~\cite{minnen2018image} first utilized a non-learnable dictionary from K-means++ algorithm \cite{arthur2007k} to improve entropy model.
In addition, Kim \textit{et al.}~\cite{kim2022joint} utilized eight learnable tokens that need to be transmitted to capture global internal dependencies.
Since its tokens are derived from the image itself and a limited number of tokens need to be transmitted, this restricts its ability to improve the entropy model.
In this paper, we propose a Dictionary-based Cross Attention Entropy model, where we employ learnable shared network parameters as the dictionary for encoding and decoding to capture external dependencies, allowing us to perform more accurate distribution estimation.

%% file: 03_method.tex
\section{Method}
\label{sec:method}

\begin{figure*}[]
    \centering
    \includegraphics[width=0.98\textwidth]{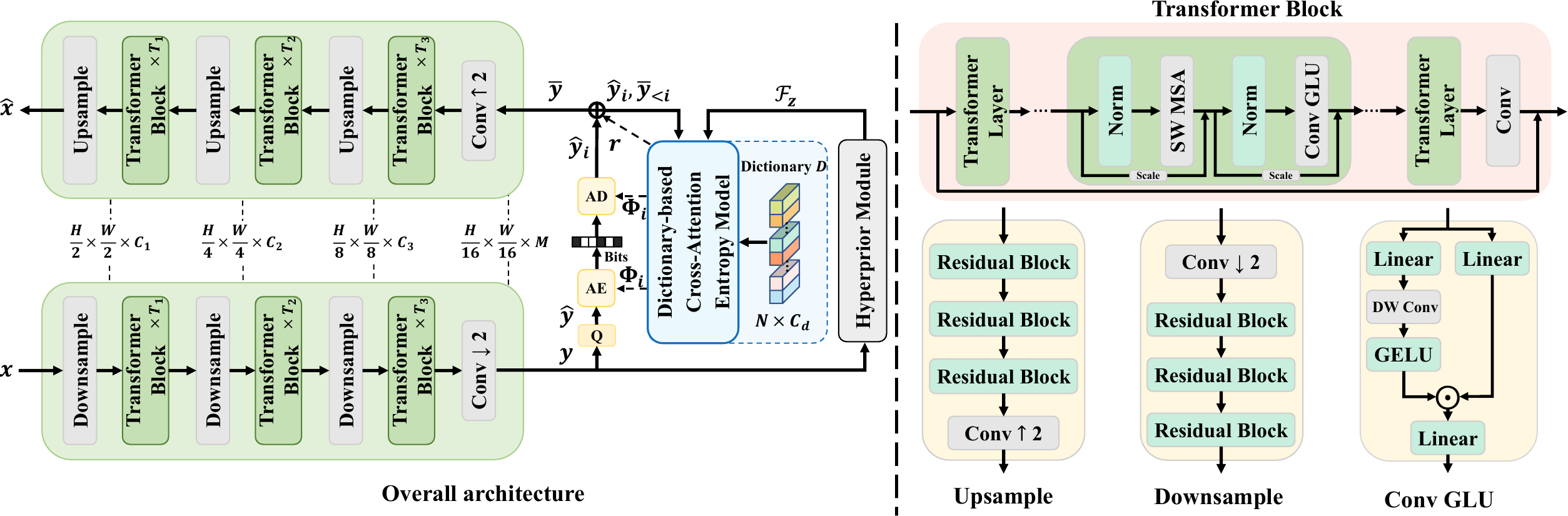}
    \caption{The overall framework of the proposed network. Given an input image $\bm{x}$, the encoder $g_a$ transforms it into the latent representation $\bm{y}$, then the proposed dictionary-based cross-attention entropy model is used to encode or decode the quantized $\hat{\bm{y}}$. Finally, the decoder $g_s$ reconstructs the image $\hat{\bm{x}}$ from the latent representation $\overline{\bm{y}}$. In the dictionary-based cross-attention entropy model, we introduce a learnable dictionary to capture typical structures and textures in natural images for improving the distribution estimation of the latent representation $\bm{y}$. 
    }
    \label{fig:framework}
\end{figure*}

\subsection{Formulation}
The overall pipeline of the learned image compression model is shown in Fig. \ref{fig:framework}.
In the encoding stage, given an input image $\bm{x}$, the encoder $g_a$ first transforms it into a latent representation $\bm{y}$: $\bm{y}=g_a(\bm{x})$.
The entropy model is then used to estimate the distribution parameters $\bm{\Phi}=\{\bm{\mu}, \bm{\sigma}\}$ of latent representation $\bm{y}$ for entropy coding.
According to previous studies \cite{minnen2018joint, he2022elic}, $\bm{y}$ is discretized to $\hat{\bm{y}} = \lceil \bm{y} - \bm{\mu} \rfloor + \bm{\mu}$ through quantization, and $\lceil \bm{y} - \bm{\mu} \rfloor$ is further losslessly encoded into bitstreams based on the evaluated distribution parameters $\bm{\Phi}$.
In the decoding stage, after restoring $\hat{\bm{y}}$ from the bitstreams, the decoder $g_s$ reconstructs the high-quality image $\hat{\bm{x}}$ from the quantized latent representation $\hat{\bm{y}}$: $\hat{\bm{x}}=g_s(\hat{\bm{y}})$.

In this process, the entropy model plays one of the most important roles, which determines the encoding length of the latent representation $\bm{y}$. 
Most existing entropy models employ the hyper-prior architecture~\cite{balle2018variational} and the channel-wise auto-regressive architecture~\cite{minnen2020channel}.
For the hyper-prior architecture, a side information $\bm{z}=h_a(\bm{y})$ is first introduced to capture the internal spatial dependencies in the latent representation $\bm{y}$.
Then, the hyper-prior decoder $h_s$ maps $\hat{\bm{z}}$ to latent feature $\bm{\mathcal{F}}_{z}$ for estimating the distribution $\{\bm{\mu}, \bm{\sigma}\} $  of latent representation $\bm{y}$: $\bm{\mathcal{F}}_{z}=h_s(\hat{\bm{z}})$.
Conditioned on $\hat{\bm{z}}$, the latent representation $\bm{y}$ is modeled as a joint Gaussian distribution:
$p_{\hat{\bm{y}}|\hat{\bm{z}}}(\hat{\bm{y}} | \hat{\bm{z}}) = \left[\mathcal{N}(\bm{\mu}, \bm{\sigma}^2) * \mathcal{U}(-\frac{1}{2}, \frac{1}{2})\right](\hat{\bm{y}})$.
For channel-wise auto-regressive architecture, $\bm{y}$ is first divided into several even slices $\{\bm{y}_0, \bm{y}_1, ..., \bm{y}_{s-1}\}$ along the channel dimension.
These slices are then encoded and decoded in sequential order.
The decoded slices $\overline{\bm{y}}_{<i} = \{ \bm{y}_0, \bm{y}_1, ..., \bm{y}_{i-1} \} $ are used to supplement information for encoding or decoding the subsequent slice $\bm{y}_i$, due to their similarity relationships.
After obtaining the hyper-prior feature $\bm{\mathcal{F}}_{z}$ and the decoded slices $\overline{\bm{y}}_{<i}$, they are entered into the entropy module $f_{E}$ to estimate $\bm{\mu}_i$ and $\bm{\sigma}_i$ of $\hat{\bm{y}}_i$ for further encoding and decoding. 
Moreover, in order to compensate for the information loss caused by quantization, $\bm{\mathcal{F}}_{z}$, $\overline{\bm{y}}_{<i}$, and $\hat{\bm{y}}_i$ are leveraged to predict the quantization error $\bm{r}_i = \bm{y}_i-\hat{\bm{y}}_i$ through the latent residual prediction net $f_{LRP}$ \cite{minnen2020channel}.
This process can be formulated as:
\begin{equation}
\label{eq:4}
\begin{split}
\bm{\mu}_i, \bm{\sigma}_i = f_{E}(\bm{\mathcal{F}}_{z}, \overline{\bm{y}}_{<i}), \quad
\bm{r}_i = f_{LRP}(\bm{\mathcal{F}}_{z}, \overline{\bm{y}}_{<i}, \hat{\bm{y}}_i).
\end{split}
\end{equation}
To train the learned image compression model, a Lagrangian multiplier-based rate-distortion optimization is employed as the loss function:
\begin{equation}\label{eq:3}
\begin{split}
\mathcal{L}&=\mathcal{R}(\hat{\bm{y}}) + \mathcal{R}(\hat{\bm{z}}) + \lambda \cdot \mathcal{D} (\bm{x}, \hat{\bm{x}}),
\end{split}
\end{equation}
where $\mathcal{R}(\hat{\bm{y}})$ and $ \mathcal{R}(\hat{\bm{z}})$ denote the bitrates of $\hat{\bm{y}}$ and $\hat{\bm{z}}$;
$\mathcal{D} (\bm{x}, \hat{\bm{x}})$ denotes the distortion between the input image $\bm{x}$ and reconstructed image $\hat{\bm{x}}$;
$\lambda$ controls the trade-off between rate and distortion.
\subsection{Dictionary-based Cross Attention Entropy Model}
\label{sec: Dictionary-based Auto-Regressive Entropy Model}
Existing methods adopt hyper-prior and auto-regression  frameworks to estimate the probability distribution of the latent representation $\bm{y}$.
Both types of methods essentially utilize the internal dependencies within latent representation to model probability distribution,
but neither explicitly leverages common patterns and textures in natural images as prior information for entropy estimation.

Our goal is to share a dictionary that preserves typical textures between the encoder and decoder. When estimating the distribution of latent representation, we can leverage the finite information to model the latent representation more accurately by querying the dictionary that retains complete information.
To achieve this goal, we propose a \textbf{D}ictionary-based \textbf{C}ross \textbf{A}ttention \textbf{E}ntropy model (\textbf{DCAE}) as shown in Fig. \ref{fig:entropy}.
Specifically, we first propose the Multi-Scale Features Aggregation module to obtain the multi-scale features $\bm{X}_{ms_i}$, which leverages features with different receptive fields to help capture textures at various scales, enabling more precise dictionary queries.
We then propose the Dictionary-based Cross Attention module to extract the dictionary features $\bm{\mathcal{F}}_{\textit{dict}_i}$ that utilizes learnable network parameters to constitute our dictionary $\bm{D}$, employing cross attention to dynamically query the complete information stored in the dictionary using the available partial texture information:
\begin{equation}
\label{eq:pipline}
\begin{split}
\bm{X}_{ms_{i}} &= \operatorname{MSFA}(\bm{X}_i), \\ 
\bm{\mathcal{F}}_{\textit{dict}_i} &= \operatorname{DCA}(\bm{X}_{ms_i}, \bm{D}),
\end{split}
\end{equation}
where $X_i=[\bm{\mathcal{F}}_{z},\overline{\bm{y}}_{<i}].$
Finally, different from Eq. \ref{eq:4}, which does not explicitly use prior information in natural images, we combine the dictionary feature $\bm{\mathcal{F}}_{\textit{dict}_i}$ with hyper-prior feature $\bm{\mathcal{F}}_{z}$ and channel-wise auto-regressive feature $\overline{\bm{y}}_{<i}$ and put them into the entropy module $f_{E}$ to obtain the parameters $\bm{\mu}_i$ and $\bm{\sigma}_i$ of the Gaussian distribution.
Furthermore, $\bm{\mathcal{F}}_{\textit{dict}_i}$, $\bm{\mathcal{F}}_{z}$ , $\overline{\bm{y}}_{<i}$ and $\hat{\bm{y}}_i$ are input to the latent residual prediction net $f_{LRP}$ \cite{minnen2020channel} to predict the quantization error $\bm{r}_i$:
\begin{equation}
\label{eq:8}
\begin{split}
\bm{\mu}_i, \bm{\sigma}_i = f_{E}(\bm{\mathcal{F}}_{z}, \overline{\bm{y}}_{<i}, \bm{\mathcal{F}}_{\textit{dict}_i}), \\
\bm{r}_i = f_{LRP}(\bm{\mathcal{F}}_{z}, \overline{\bm{y}}_{<i}, \bm{\mathcal{F}}_{\textit{dict}_i}, \hat{\bm{y}}_i).\\
\end{split}
\end{equation}

\paragraph{Multi-Scale Features Aggregation Module.} 
The proposed Multi-Scale Features Aggregation Module is used to capture multi-scale textures from feature maps, enabling more accurate querying of prior information stored in the dictionary.
In Multi-Scale Features Aggregation Module, we first utilize feature maps from different convolutional layers to achieve multi-scale texture extraction. 
Feature maps from shallow convolutional layers have limited receptive fields, enabling them to capture finer textures. As the convolution depth increases, deeper layers can capture larger-scale textures.
In addition, to achieve efficient convolutional computation, each basic convolution unit consists of two linear layers and a 3×3 depthwise (DW) convolution:
\begin{equation}\label{add}
\begin{split}
\operatorname{EConv}(\bm{X})&=\operatorname{DWConv_{3\times3}}(\bm{XW}^{in})\bm{W}^{out},\\
\bm{X}_{i}^{1}=\bm{X}_{i}, &\quad \bm{X}_{i}^{j}=\operatorname{EConv}(\bm{X}_{i}^{j-1}),\\
\bm{X}_i^{merge}&=[\bm{X}_i^1,...,\bm{X}_i^m]\bm{W^}{merge},\\
\end{split}
\end{equation}
where $\bm{W}^{in},\bm{W}^{out}$ are used to perform the transformation of convolution channels and $\bm{W}^{merge}$ is used to merge multi-scale information;
$m$ represents the number of stacked efficient convolutional layers.
To enable more precise queries of prior information from the dictionary, we then employ spatial attention module \cite{woo2018cbam} to dynamically assign a weight to each spatial position.
\begin{equation}
% \begin{split}
\operatorname{MSFA}(\bm{X}_i)=\operatorname{SA}(\bm{X}_i^{merge})\odot{X}_i^{merge}.
% \end{split}
\end{equation}
where $\operatorname{SA}(\bm{X}_i^{merge})\in \mathbb{R}^{H \times W}$ represents the spatial weight map output by the spatial attention module.

\paragraph{Dictionary-based Cross Attention Module.} 
After introducing how we utilize features with different receptive fields to capture textures at various scales, we will  introduce how to use the existing information to perform dictionary queries.

We first established a shared dictionary $\bm{D}$ using learnable network parameters to preserve the common textures found in natural images.
The dictionary $\bm{D}$ is initialized as a tensor with the shape of $[N, C_d]$, where $N$ is the number of dictionary entries and $C_d$ is the number of feature dimensions.
During the training process, the dictionary will gradually learn to fit the typical structures, which is similar to traditional dictionary learning methods~\cite{yang2010image, zeyde2012single} learning to fit natural images with the image patch dictionary.
In addition, it is noteworthy that this dictionary is simultaneously shared between the encoder and decoder, thereby not requiring additional bitrate for transmission.

\begin{figure*}[]\label{fig:entropy}
    \centering
    \includegraphics[width=0.9\textwidth]{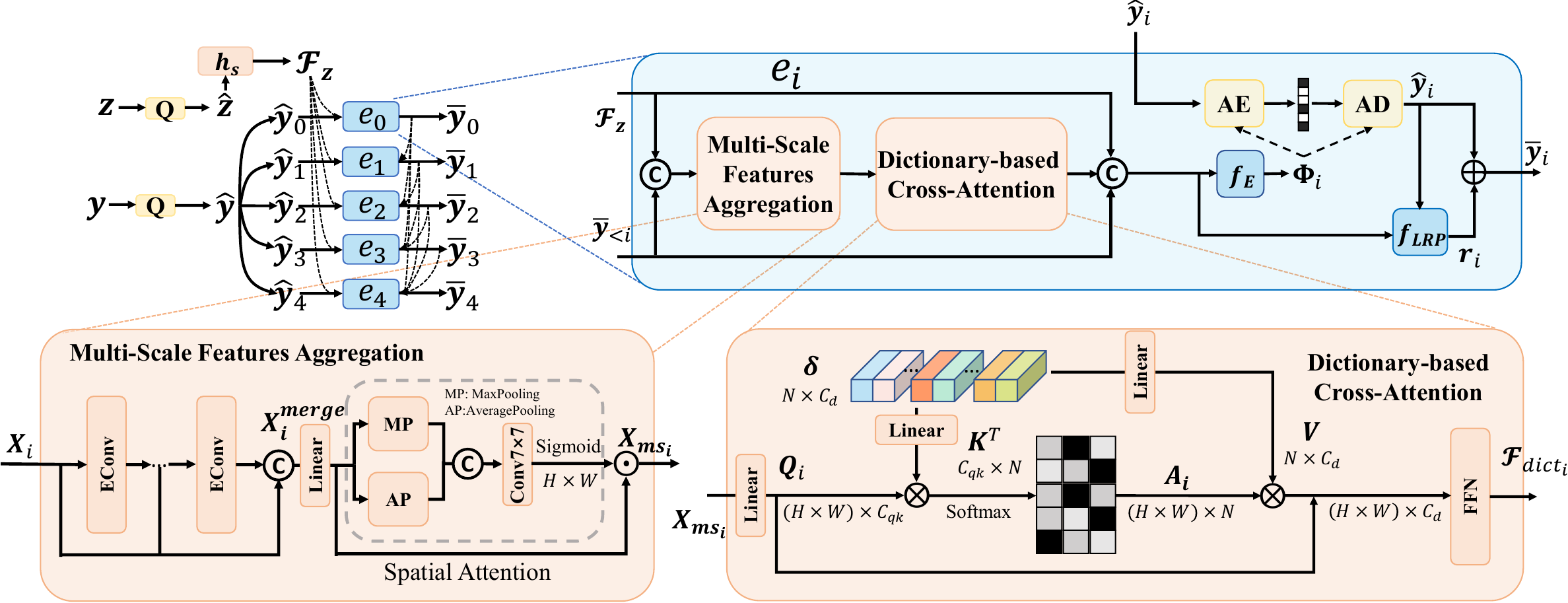}
    \caption{The proposed Dictionary-based Cross Attention Entropy model. The Dictionary-based Slice Network $e_i$ is used to encode or decode the latent representation $\hat{\bm{y}}_i$. In $e_i$, the hyper-prior feature $\bm{\mathcal{F}}_{z}$ and channel-wise auto-regressive feature $\overline{\bm{y}}_{<i}$ are first fed into our Multi-Scale Features Aggregation module to obtain multi-scale features $\bm{X}_{ms_i}$.
    Then, the multi-scale features $\bm{X}_{ms_i}$ are used to query the dictionary to extract the dictionary feature $\bm{\mathcal{F}}_{\textit{dict}_i}$. Finally, the dictionary feature $\bm{\mathcal{F}}_{\textit{dict}_i}$ is taken as input to the entropy module $f_E$ to estimate the distribution parameters $\Phi_i$ of $\hat{\bm{y}}_i$ for entropy coding, and to the latent residual prediction net $f_{LRP}$ to predict the quantization error $\bm{r}_i$.  }
    \label{fig:entropy}
\end{figure*}
We then apply cross attention to query the learnable dictionary with features that contain partial texture information, aiming to capture prior information present in natural images.
Specifically, we use the multi-scale feature $\bm{X}_{ms_i}$ to generate the query tokens $\bm{Q}_i$, which contains partial texture information.
The learnable dictionary $\bm{D}$ is used to generate the key tokens $\bm{K}$  and the value tokens $\bm{V}$.
The key tokens $\bm{K}$ are utilized for calculating similarity with the query tokens, while the value tokens $\bm{V}$ represent the texture information stored in the dictionary.:
\begin{equation}\label{eq:5}
\bm{Q}_i = \bm{X}_{ms_i} \bm{W}^Q, \  \bm{K} = \bm{D} \bm{W}^K, \ \bm{V} = \bm{D},
\end{equation}
where $\bm{W}^{Q} \in \mathbb{R}^{ C_{ms} \times C_{qk}}$, $\bm{W}^K \in \mathbb{R}^{C_d \times C_{qk}} $ are linear transforms for $\bm{X}_{ms}$ and $\bm{D}$, respectively.
Then, we can obtain the dictionary feature $\bm{\mathcal{F}}_{\textit{dict}}$ in a cross-attention manner:
\begin{equation}
\label{eq:6}
\bm{A}_i = \operatorname{SoftMax} ( \bm{Q}_i\bm{K}^T/\bm{\tau} ), \quad 
\bm{\mathcal{F}}_{\textit{dict}_i} = \bm{A}_i  \bm{V},
\end{equation}
where $\bm{\tau}_i$ is a learnable scaling parameter to adjust the range of the dot product of $\bm{Q}_i$ and $\bm{K}$.
$\bm{Q}_i\bm{K}^T \in \mathbb{R}^{(H \times W) \times N}$ represents the similarity map between query feature and dictionary entries.
The softmax function is applied along the dimension 
 $N$ to normalize the weights of each dictionary entry.
Finally, we utilize the normalized weights $\bm{A}_i$ to perform a weighted aggregation $\bm{A}_i \bm{V}$ of the dictionary information, resulting in the dictionary feature $\bm{\mathcal{F}}_{\textit{dict}_i}$ and the $\bm{\mathcal{F}}_{\textit{dict}_i}$ is further enhanced by FFN layer.
\subsection{Network Architecture}
Our main encoder $g_a$ and decoder $g_s$ is designed as Fig. \ref{fig:framework} shows.
We apply basic downsampling/upsampling modules followed by Swin transformer blocks \cite{liang2021swinir} to perform non-linear mapping for extracting compact latent representation.
Each basic downsampling / upsampling module consists of a strided / transposed convolution, along with several cascaded residual blocks to extract the local context information.
The Swin transformer blocks are used to capture long-range dependencies, supplying non-local information.
In addition, in order to further enhance  non-linear transformation capability of our transformer, inspired by \cite{yu2023metaformer} and \cite{shi2024transnext}, we employ ResScale \cite{yu2023metaformer} to scale up transformer model size from depth and we adopt Convolutional Gated Linear Unit \cite{shi2024transnext} to construct our FFN.

Previous methods \cite{liu2023learned, fu2024weconvene, he2022elic}, typically use the same number of channels and modules across different stages of the encoder-decoder. However, computing high-resolution features slows down the overall model speed. To achieve a more efficient structural design, we adopt varying numbers of channels and modules at different stages, shifting computations to lower-resolution stages to enable faster encoding and decoding speed. 
A more detailed description of this design can be found in Section \ref{sec:4.1}.

%% file: 04_experiments.tex
\section{Experiments}
\label{sec:experiments}
\begin{table*}[htbp]
    \footnotesize
    \centering
    \caption{Computation burden and performance comparisons between different methods.}
    \vspace{0.2cm}
    \label{tab:compare}
    \setlength\tabcolsep{3pt} 
    \begin{tabular}{p{3.5cm}|>{\centering\arraybackslash}p{0.8cm}>{\centering\arraybackslash}p{0.8cm}>{\centering\arraybackslash}p{0.8cm}>{\centering\arraybackslash}p{0.8cm}>{\centering\arraybackslash}p{0.8cm}>{\centering\arraybackslash}p{1.1cm}|>{\centering\arraybackslash}p{1.1cm}>{\centering\arraybackslash}p{1.1cm}>{\centering\arraybackslash}p{1.1cm}}
    \toprule
    \multirow{2}{*}{\textbf{Model}} & \multicolumn{3}{c}{\textbf{Latency (ms)}} & \multicolumn{2}{c}{\textbf{GFLOPs}} & \multirow{2}{*}{\textbf{Params}} & \multicolumn{3}{c}{\textbf{BD-rate}} \\
     & \textbf{Tot.} & \textbf{Enc.} & \textbf{Dec.} & \textbf{Enc.} & \textbf{Dec.} & & \textbf{Kodak} & \textbf{Tecnick} & \textbf{CLIC} \\
    \midrule
    STF (CVPR'22) \cite{zou2022devil}  & 233 & 102 & 131 & 143 & \textbf{161} & 99.8M & -4.3\% & - & -4.1\%  \\
    WACNN (CVPR'22) \cite{zou2022devil}  & 193 & \textbf{80} & 113 & 138 & 231 & 75.0M & -4.8\% & - & -4.4\%  \\
    ELIC (CVPR'22) \cite{he2022elic}  & 210 & 91 & 119 & 138 & 233 & \textbf{41.9M} & -7.1\% & - & -  \\
    % GLLMM  & - & - & - & - & - & - & -3.3\% & -5.5\% & -  \\
    M2T (ICCV'23) \cite{mentzer2023m2t}  & - & - & - & - & - & - & -8.5\% & - & -  \\
    MT (ICCV'23) \cite{mentzer2023m2t}  & - & - & - & - & - & - & -12.5\% & - & -  \\
    TCM (CVPR'23) \cite{liu2023learned}  & 293 & 142 & 151 & 307 & 441 & 75.9M & -11.8\% & -12.0\% & -12.0\%  \\
    % MLIC (ACMMM'23) \cite{jiang2023mlic}  & - & - & - & - & - & - & -9.8\% & -13.6\% & -12.8\%  \\
    MLIC+ (ACMMM'23) \cite{jiang2023mlic}  & - & - & - & - & - & - & -13.1\% & -17.3\% & -16.4\%  \\
    MLIC++ (NCW ICML'23) \cite{jiang2023mlic} & 772 & 362 & 410 & 222 & 300 & 116.5M & -15.1\% & -18.6\% & -16.9\%  \\
    FTIC (ICLR'24) \cite{li2023frequency} & - & - & - & \textbf{127} & 355 & 71.0M & -14.6\% & -15.1\% & -13.6\%  \\
    WeConvene (ECCV'24) \cite{fu2024weconvene}  & 545 & 275 & 271 & 702 & 320 & 105.5M & -8.5\% & -9.2\% & -10.1\%  \\
    CCA (NeurIPS'24) \cite{han2024causal}  & 223 & 122 & 101 & 277 & 394 & 64.9M & -13.7\% & -15.3\% & -14.5\%   \\
    Ours & \textbf{193} & 93 & \textbf{100} & 252 & 305 & 119.2M & \textbf{-17.0\%} & \textbf{-21.1\%} & \textbf{-19.7\%}  \\
    \midrule
    VVC & - & - & - & - & - & - & 0\% & 0\% & 0\% \\
    \bottomrule
    \end{tabular}
\end{table*}
\begin{figure*}[htbp]
    \centering
    \begin{minipage}{0.32\linewidth}
        \centering
        \includegraphics[width=\linewidth]{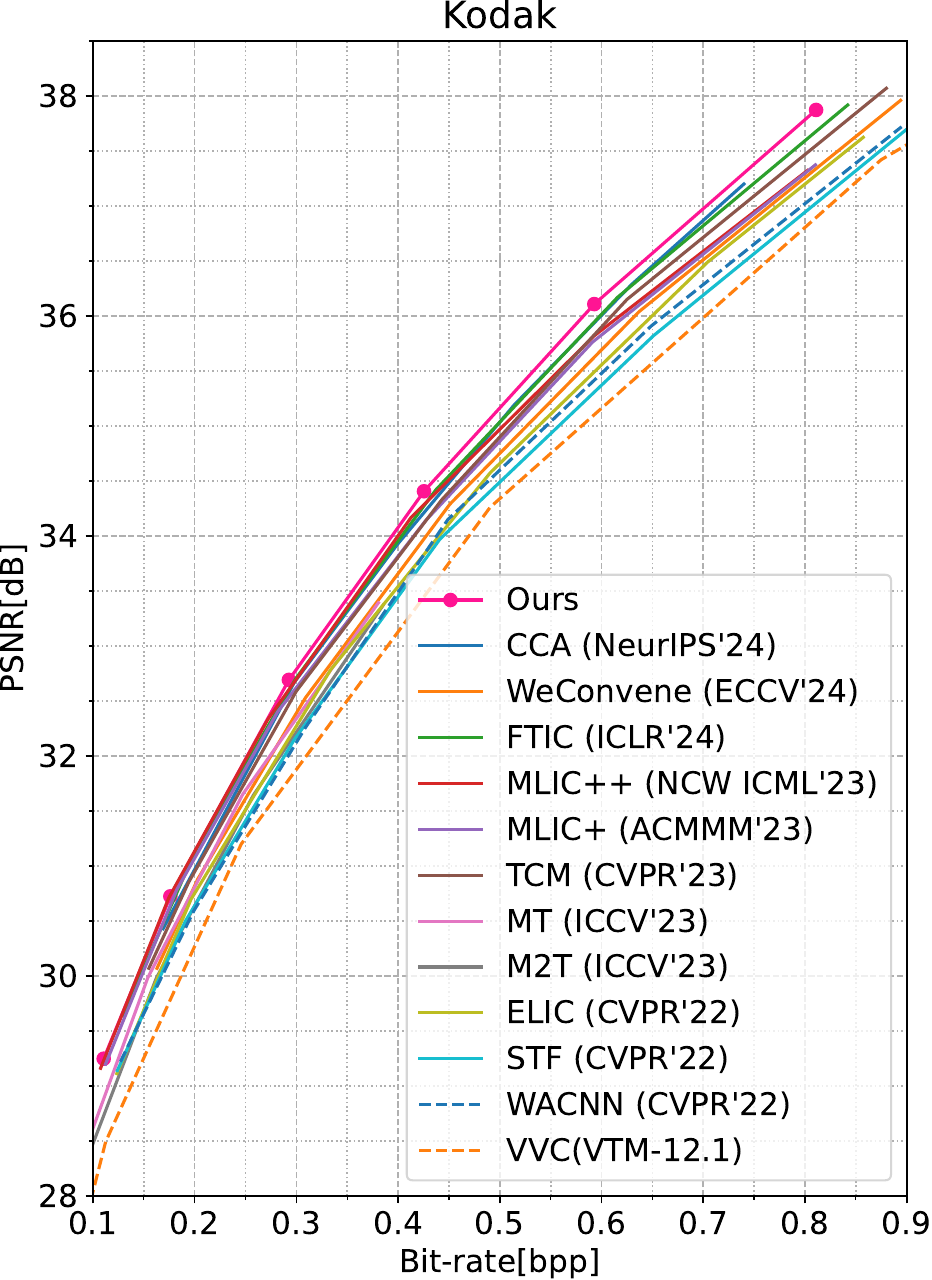}
        \caption{Performance evaluation (PSNR) on the Kodak dataset.}
        \label{fig:Kodak}
    \end{minipage}
    % \hfill
    \begin{minipage}{0.32\linewidth}
        \centering
        \includegraphics[width=\linewidth]{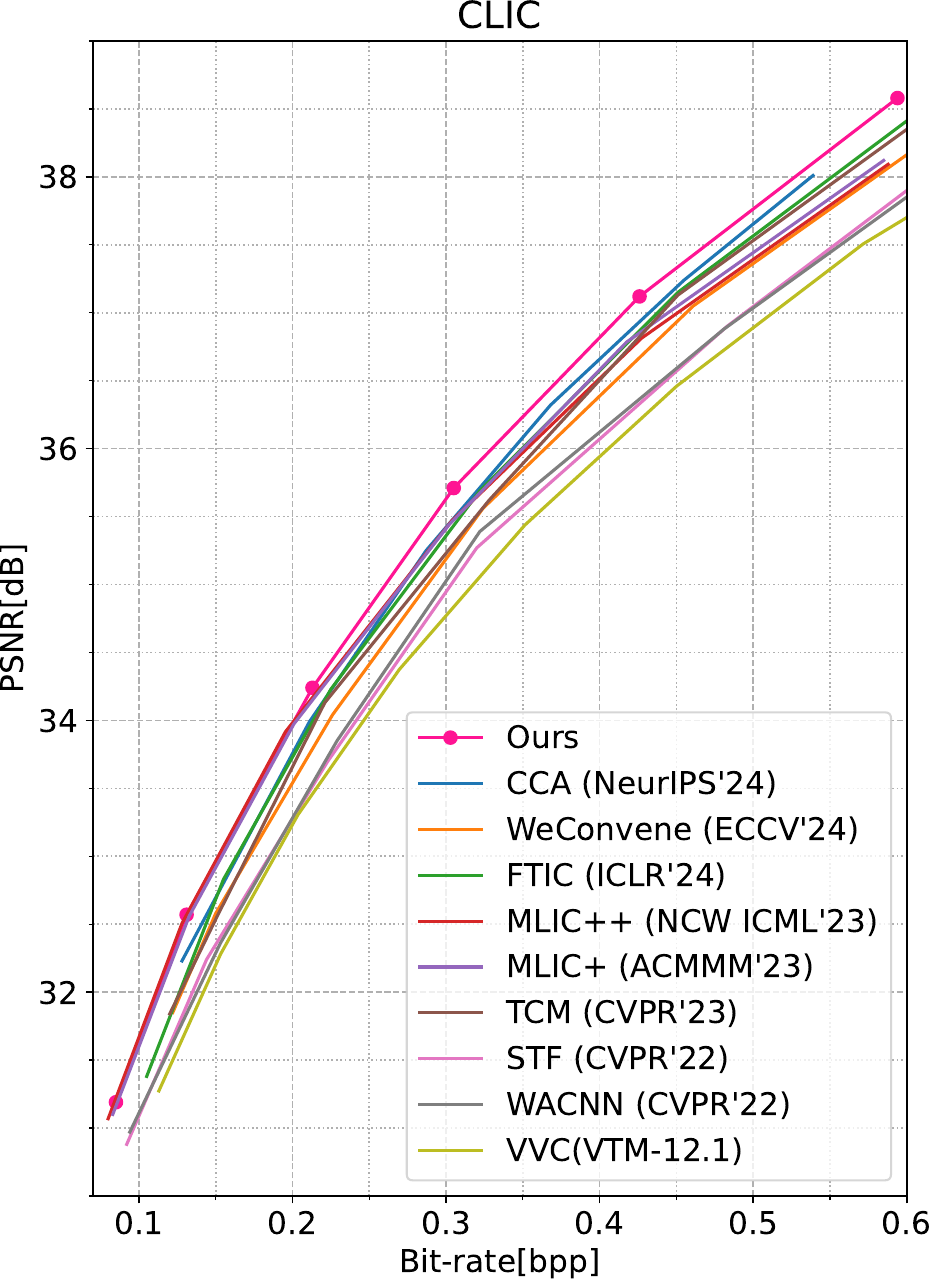}
        \caption{Performance evaluation (PSNR) on the CLIC dataset.}
        \label{fig:CLIC} 
    \end{minipage}
    % \hfill
    \begin{minipage}{0.32\linewidth}
        \centering
        \includegraphics[width=\linewidth]{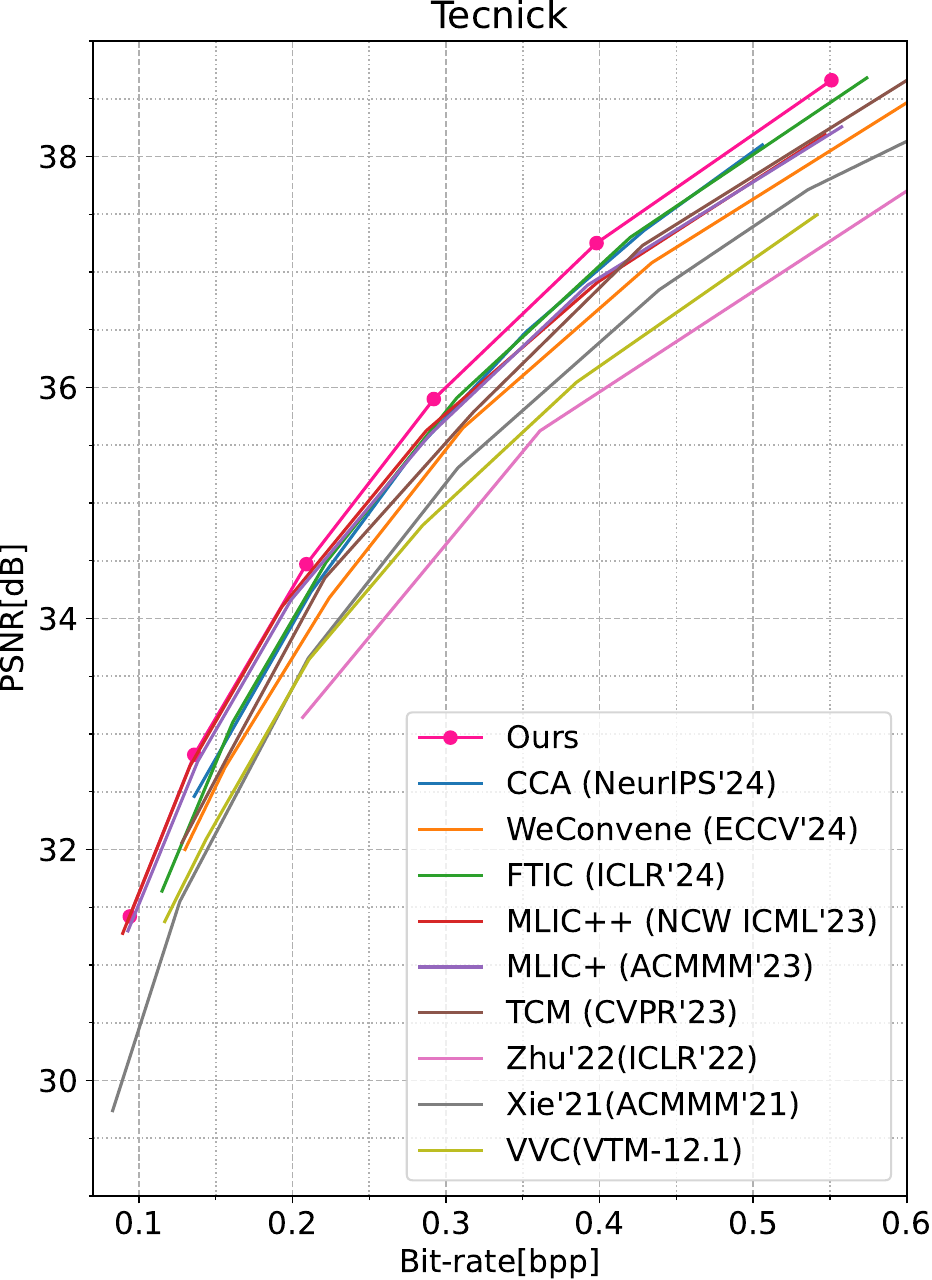}
        \caption{Performance evaluation (PSNR) on the Tecnick dataset.}
        \label{fig:Tecnick}
    \end{minipage}
\end{figure*}
\subsection{Experimental Settings}
\label{sec:4.1}
\paragraph{Training Details.}
We follow the experimental settings of recent state-of-the-art methods \cite{zou2022devil} and utilize the OpenImages dataset \cite{Krasin2016openimage} to train our final model.
OpenImages dataset contains 300k images with short edge no less than 256 pixels. 
We randomly crop patches of size 256 × 256 for each training iteration, with a batch size of 16, and adopt the Adam optimizer \cite{kingma2014adam} to minimize the R-D loss in Eq.\ref{eq:3}.
We utilize Mean Squared Error (MSE) loss as the distortion measure to train our models.
In order to obtain compression models with different compression ratio, we train our models with different $\lambda$ values, \textit{i.e.}, $\lambda=\{0.0018, 0.0035, 0.0067, 0.0130, 0.0250, 0.0500\}$.
We train our models with a initial learning rate $1e-4$ for 80 epochs (1.5 million iterations), and then decrease the learning rate to $1e-5$ and train the models for another 20 epochs (0.375 million iterations) for obtaining the final models. 
We use RTX 4090 to complete our experiments.

\paragraph{Implementation Details.}
For our model, we introduce a learnable dictionary with 128 dictionary entries and 640 channels.
We set the number of transformer layers within Transformer Block at different scales as $(T_1, T_2, T_3) = (1, 2, 12)$ for our model; where in the hyper-prior module, our model contain 1 transformer layer.
As for the feature dimension, we set the feature dimension for different scales as $(C_1, C_2, C_3) = (96, 144, 256)$.
The dimensions of the latent representation $\bm{y}$ and side information $\bm{z}$ are set to 320 and 192, respectively.
The head dimensions of transformer layers in encoder $g_a$ and decoder $g_s$ are set as \{8, 16, 32, 32, 16, 8\}, while the head dimensions of transformer layers in hyper-prior encoder $h_a$, hyper-prior decoder $h_s$ and Dictionary-based Cross Attention are set 32.
The window sizes of these transformer layers are set as  8$\times$8 for encoder $g_a$ and decoder $g_s$ and 4$\times$4 for the hyper-prior module.

\paragraph{Comparison Methods and Benchmark Datasets.}
We choose three benchmark datasets, \textit{i.e.}, Kodak image set \cite{kodak}, Tecnick testset \cite{Asuni_Giachetti_2014}, CLIC professional validation dataset \cite{clic} to evaluate our methods.
The competing appraches including classical standard VVC (VTM-12.1)~\cite{Rao_Dominguez_2022} and recent state-of-the-art LIC models \cite{xie2021enhanced, zou2022devil, he2022elic, liu2023learned, zhu2021transformer, cheng2020learned,mentzer2023m2t,li2023frequency, jiang2023mlic, jiang2023mlicpp, fu2024weconvene, han2024causal}.
VVC results are achieved by CompressAI~\cite{begaint2020compressai}, while the results of other methods are provided by the method authors.

\subsection{Ablation Studies}
In this part, we conduct experiments to validate the effectiveness of the proposed Dictionary-based Cross-Attention Entropy (DCAE) model.
We use a smaller model for ablation studies, where the number of transformer layers is set to $(T_1, T_2, T_3) = (0, 0, 4)$.
All ablation studies are trained with a initial learning rate $1e-4$ for 20 epochs (0.75 million iterations) with a batch size of 8, followed by 5 epochs (0.1875 million iterations) of training with a learning rate of $1e-5$.
\begin{figure*}[]
    \centering
    \includegraphics[width=0.90\textwidth]{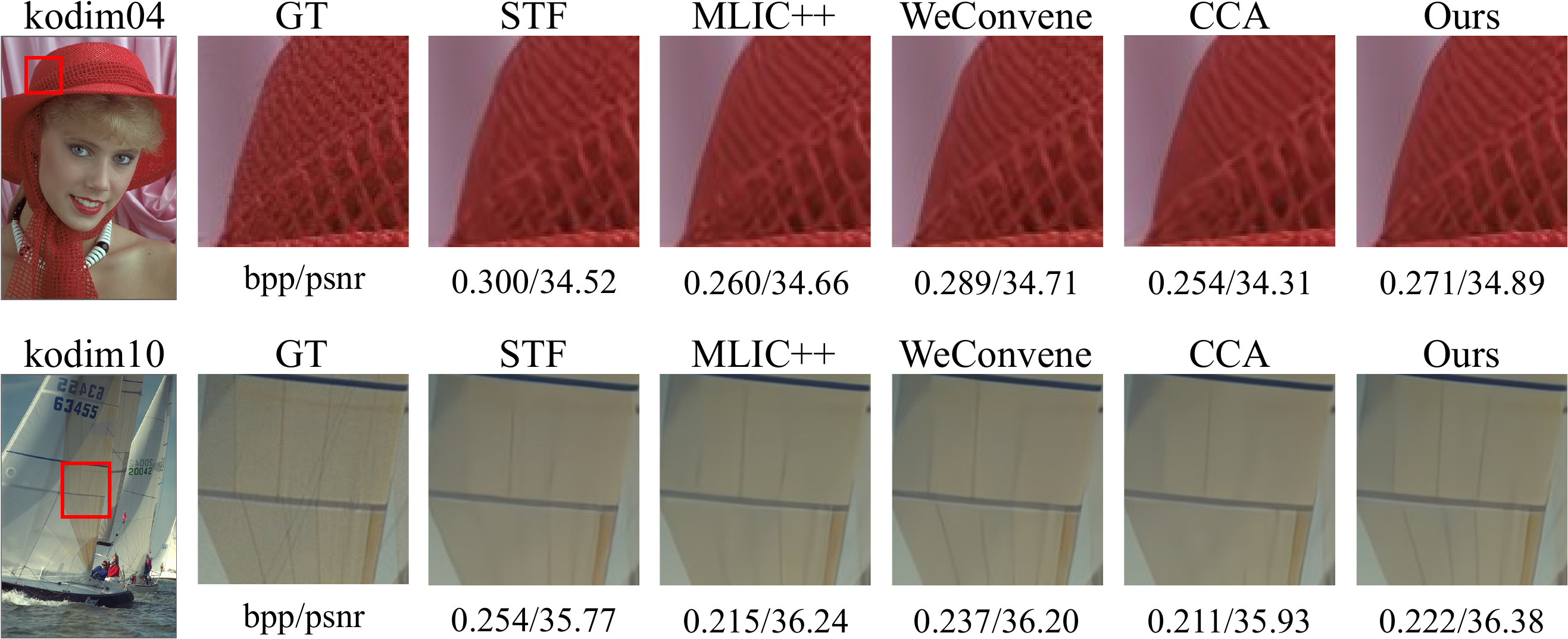}
    \caption{Reconstructed images \textit{kodim04} and \textit{kodim10} from the Kodak dataset. In the above visualization, our model effectively restores the texture information while maintaining comparable or lower bitrate.}
    \label{fig:example}
    \vspace{-10pt}
\end{figure*}
\setlength{\tabcolsep}{17pt} 
\begin{table}[ht]
    \vspace{-1pt} % Adjust vertical space as needed
        \centering
        % Right content: the table
        \caption{Ablation studies of the proposed modules.}
        \begin{tabular}{lcc}
        \toprule
         Model & BD-rate & Latency (ms)\\
        \midrule
        baseline  &  -4.20\% & 143\\
        + DCA & -7.28\% & 153 \\
        + MSFA & -8.50\% & 160 \\
        % \midrule
        % VVC & 0\% & -\\
        \bottomrule
        \end{tabular}
        \label{tab:modules ablation}
    \vspace{-15pt}
    % \vspace{-10pt} % Adjust vertical space as needed
\end{table}
\paragraph{Effects of DCA and MSFA.}
In order to show the effectiveness of the proposed methods, we remove all proposed modules to establish a baseline and progressively add them back to demonstrate the benefits they provide.
% \vspace{-5pt}
% \vspace{-10cm}
As shown in Tab.\ref{tab:modules ablation}, the DCA achieves a 3.08\% improvement in BD-rate on kodak dataset while only increasing the coding latency of baseline model from 143 ms to 153 ms.
Furthermore, MSFA further improves the BD rate from 7.28\% to -8.50\%.
\setlength{\tabcolsep}{2pt}  
\begin{table}[ht]
    \centering
    \caption{Ablation studies of the dictionary size.}
    \begin{tabular}{lcccccc}
    \toprule
    $N$     & - & 64 & 128 & 192 & 256 \\
    \midrule
    BD-rate & -4.20\% & -6.84\% & \textbf{-7.28\%} & -7.26\% & -6.92\% \\
    \midrule
    Latency (ms) & 143 & 153 & 153 & 154 & 154 \\
    \bottomrule
    \end{tabular}
    \label{tab:dictionary size ablation}
    \vspace{-15pt}
\end{table}
\paragraph{Effects of dictionary size.}
In order to analyze the effect of dictionary size, we also train models with 
64, 192 and 256  dictionary items, respectively.
The BD-rate values on the Kodak dataset by different models are reported in 
Tab.\ref{tab:dictionary size ablation}.
When equipped with the DCA module, even with a smaller number of dictionary entries, \textit{i.e.}, 64, the performance improves significantly, with the BD-rate improving from -4.20\%  to -6.84\%.
As the number of dictionary entries further increases to 128, the performance of the model improves further, reaching -7.28\%.
However, when the number of dictionary entries increases again, the improvement become saturated and does not result in better compression performance.
\begin{table}[ht]
    \centering
    \caption{Ablation studies of the MSFA.}
    \begin{tabular}{lcccccc}
    \toprule
    $m$  & 0 & 1 & 2 & 3 & 4 \\
    \midrule
    BD-rate & -7.28\% & -7.62\% & -8.04\% & \textbf{-8.50\%} & -8.36\% \\
    \midrule
    Latency (ms) & 153 & 157 & 158 & 160 & 162 \\
    \bottomrule
    \end{tabular}
    \label{tab:MSFA ablation}
    \vspace{-15pt}
\end{table}
\paragraph{Effects of the number of convolutional layers in MSFA.}
Increasing the number of convolutional layers $m$ in MSFA will aid in modeling multi-scale features and enlarging the receptive field, thereby facilitating accurate dictionary queries.
Tab.\ref{tab:MSFA ablation} shows the impact of the number of convolutional layers.
Increasing the number of convolutional layers $m$ from 1 to 3 improves the  performance of our model, achieving a peak of -8.50\% when $m$ is 3.
Further increasing $m$  does not lead to any additional performance gains.

\paragraph{Comparison with Global Token.}
Kim \textit{et al.}~\cite{kim2022joint} proposed the global token to capture the global internal dependencies of latent representation.
Both the global token and our dictionary use learnable network parameters to improve the entropy model.
However, since the global token must generate distinct tokens for each image, it necessitates the transmission of these tokens during both encoding and decoding processes. 
In contrast, our dictionary captures common textures across different images, enabling its shared usage between the encoder and decoder.
In addition, since the global token utilizes a relatively small number of tokens, whereas our dictionary employs a significantly larger number of dictionary entries (128 vs. 8), our dictionary can demonstrate superior representational capacity.
To ensure a fair comparison, we apply the global token to our baseline, with the results presented in Tab.\ref{tab:global token}.
It can be observed that, under comparable latency (153 vs. 152), our DCA achieves better performance (-7.28\% vs. -6.59\%).
% \vspace{-5pt}
\setlength{\tabcolsep}{15pt}  
\begin{table}[ht]
    \vspace{-1pt} % Adjust vertical space as needed
        \centering
        % Right content: the table
        \caption{Comparison with global token.}
        \begin{tabular}{lcc}
        \toprule
         Model & BD-rate & Latency (ms)\\
        \midrule
        baseline  &  -4.20\% & 143\\
        DCA & -7.28\% & 153 \\
        global token & -6.59\% & 152 \\
        % \midrule
        % VVC & 0\% & -\\
        \bottomrule
        \end{tabular}
        \label{tab:global token}
\end{table}
\subsection{Comparisons with State-of-the-Art Methods}
The rate-distortion performance by different methods on Kodak dataset, CLIC dataset, and Tecnick dataset are shown in Fig. \ref{fig:Kodak}, Fig. \ref{fig:CLIC}, and Fig. \ref{fig:Tecnick}, which use PSNR to evaluate performance.
Our proposed method consistently outperform the existing methods on all the three benchmark datasets.
Additionally, we present the BD-rate results, GFLOPs and the compression latency information by our method and the current state-of-the-art methods
in Tab. \ref{tab:compare}.
The RD-rate \cite{Bjontegaard_2001} value is calculated with VVC (VTM-12.1) as the anchor.
The latency and GFLOPs are calculated on the Kodak dataset.
As can be found in the table, compared to MLIC++,
which currently achieves the best BD-rate performance, our model outperforms it across all three datasets.
Notably, the latency of MLIC++ on Kodak dataset is nearly four times that of our model.
A more detailed Rate-speed comparison can be found in Fig. \ref{fig:latency_bd_rate}, which clearly demonstrates that our model is able to achieve good compression results with a smaller latency.
Some visual examples by our proposed model as well as recent state-of-the-art methods are shown in Fig. \ref{fig:example}.
The visual results clearly validate the superiority of our model in keeping image details.
\begin{figure}[]
    \label{fig:2}
    \centering
    \includegraphics[width=0.45\textwidth]{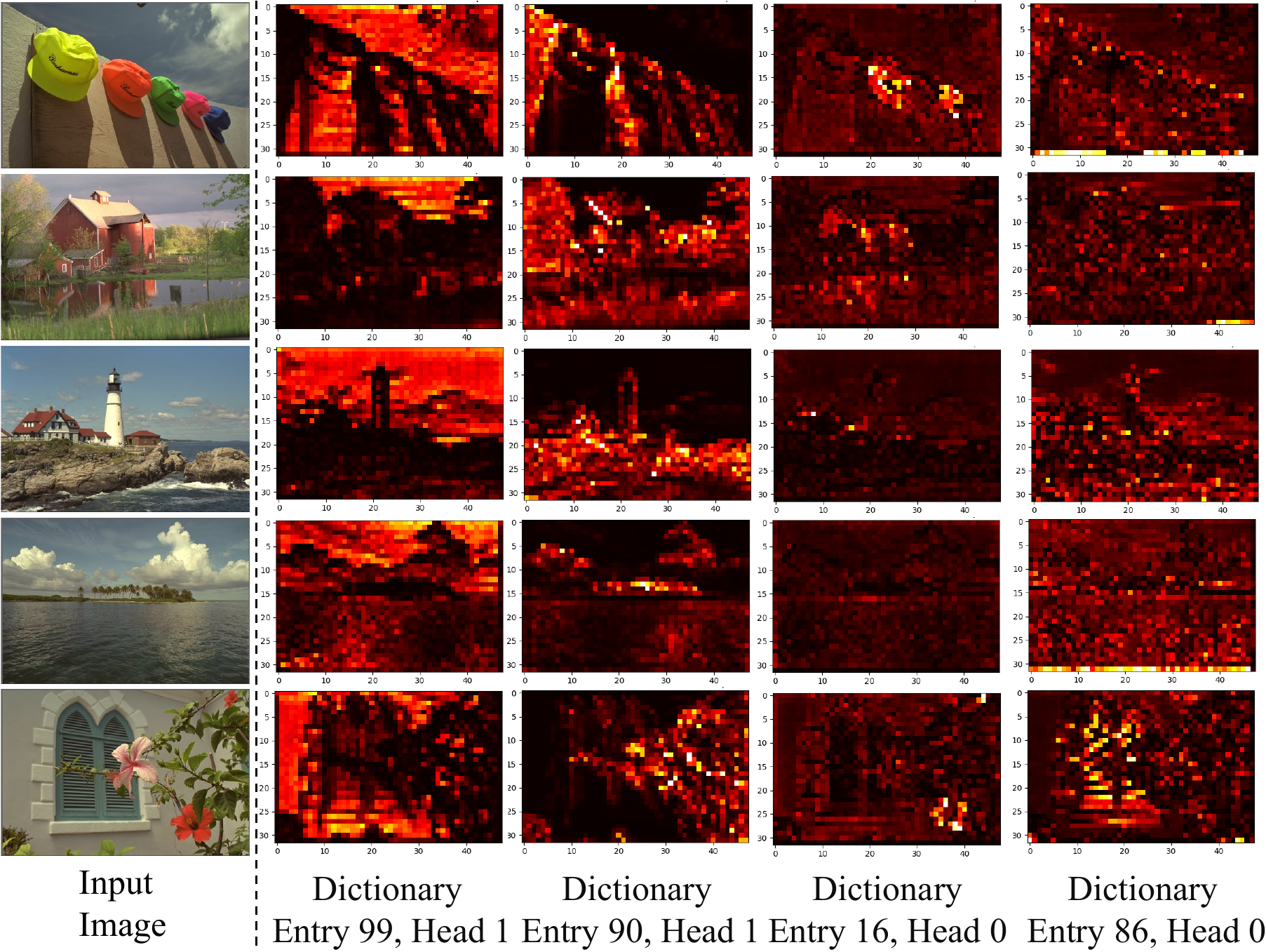}
    \caption{ Visualization of attention maps between feature maps and dictionary entries. The first column represents original images from the Kodak dataset and the last four columns represent attention maps of a specific dictionary entry across different images.}
    % \vspace{-12pt}
    \label{fig:2}
\end{figure}
\subsection{Visualization Analysis}
In order to analyze our argument of leveraging typical local structures, we present some intermediate attention maps to analyze the behaviour of our model.
In Fig. \ref{fig:2}, we present the same attention maps on different testing images.
We can clearly observe corelations between dictionary items and image local structures, similar image local structures tend to leverage the same dictionary item to predict the latent representation.
The results validate our idea of exploiting dictionary to provide prior information of typical structures.

%% file: 10_conclusion.tex
\section{Conclusion}
\label{sec:conclusion}
In this paper, we propose a novel dictionary-based cross-attention entropy (DCAE) model for explicitly capturing prior information from the training dataset. The proposed entropy model uses learnable network parameters to summarize the typical structures and textures in natural images, thereby improving the entropy model. We show that DCAE brings effective improvement in RD performance. By incorporating the proposed DCAE, we exceed the state-of-the-art RD performance on three different resolution datasets (\textit{i.e.}, Kodak, Tecnick, CLIC Professional Validation).

%% file: 05_acknowledgment.tex
\section{Acknowledgment}
\label{sec:acknowledgment}
This work was supported by National Natural Science Foundation of China (No. 62476051, No. 62176047) and Sichuan Natural Science Foundation (No. 2024NSFTD0041).

%% file: 12_appendix.tex
\maketitlesupplementary
\section{More detailed network architecture}
\label{sec:More detailed network architecture}
The overall framework of the proposed network. The encoder-decoder and hyperprior module are composed of Transformer blocks, downsample modules, upsample modules, and convolutions.
We set the number of Transformer layers to 1 in the hyperprior module and we use the Factorized Entropy Model to estimate the distribution of sid\begin{figure*}[ht]
    \centering
    \includegraphics[width=0.95\textwidth]{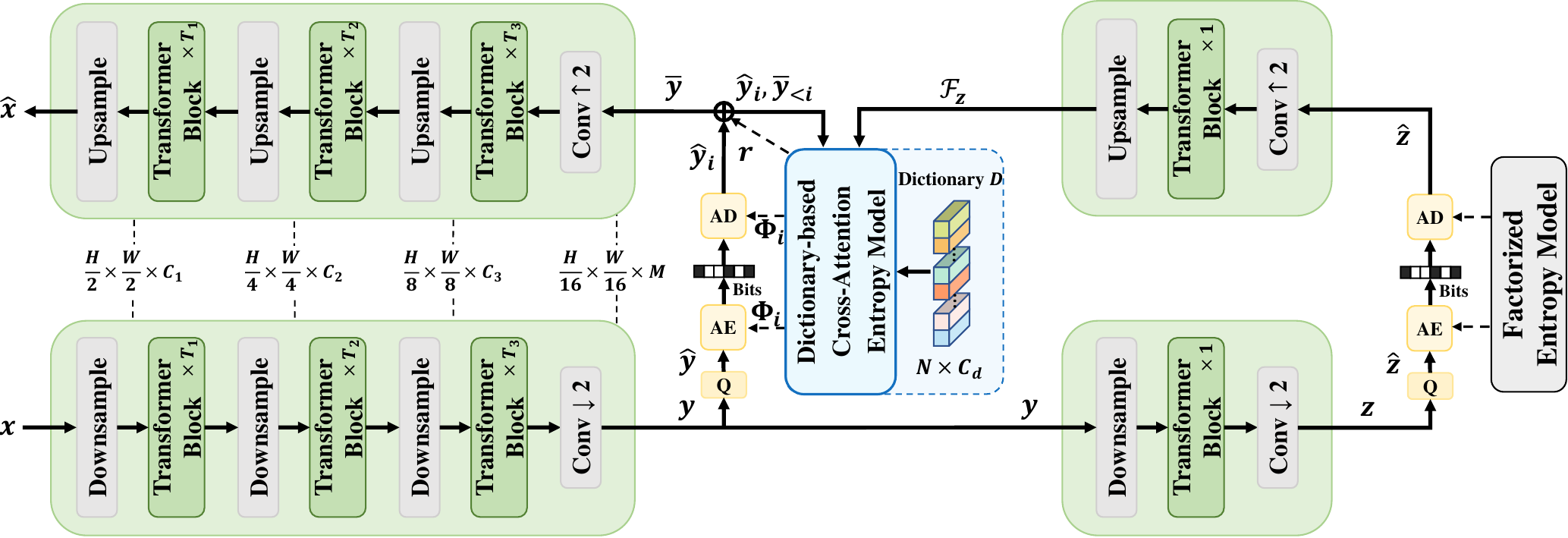}
    \caption{The overall framework of the proposed network.}
    \label{fig:example_a}
\end{figure*}
\vspace{-2mm}
\section{More ablation studies}
To further demonstrate the effectiveness of our entropy model, we replace the entropy models in three state-of-the-art methods, ELIC (CVPR'22), TCM (CVPR'23), and FTIC (ICLR'24), with ours. All studies are trained with an initial learning rate of $1e-4$ for 20 epochs, followed by 5 epochs at $1e-4$ , using a batch size of 8.
As shown in Tab. \ref{tab: Comparative evaluation}, ELIC's BD-rate improves by 3.75\% (-2.62\% to -6.37\%), TCM by 2.41\% (-7.80\% to -10.21\%), and FTIC by 3.19\% (-6.52\% to -9.71\%). In addition, our entropy model can enhance the existing methods in an plug-and-play manner. Furthermore, combining our model with ELIC's further improves BD-rate from -2.62\% to -9.66\% (ELIC uses a spatial-channel autoregressive model, while ours is channel-wise only).
\begin{table*}[h]
    \centering
    \caption{Comparative evaluation by replacing the entropy model.}
    \small 
    \resizebox{0.95\textwidth}{!}{
    \begin{tabular}{lccccccc}
    \toprule
    Autoencoder & \multicolumn{3}{c}{ELIC} & \multicolumn{2}{c}{TCM} & \multicolumn{2}{c}{FTIC} \\
    \cmidrule(lr){2-4} \cmidrule(lr){5-6} \cmidrule(lr){7-8}
    Entropy Model & ELIC & ours & ELIC+ours & TCM & ours & FTIC & ours \\
    \midrule
    BD-rate & -2.62\% & -6.37\% & -9.66\% & -7.80\% & -10.21\% & -6.52\% & -9.71\% \\
    \bottomrule
    \end{tabular}
    }
    \label{tab: Comparative evaluation}
\end{table*}

\section{Multi-Scale visualizations}
To study how multi-scale context influences dictionary query, we visualize attention maps at different feature levels from the EConv layers. Specifically, we use three EConv layers for feature extraction and progressively set each layer's output to zero, starting from the last, creating four models: ``scale4'', ``scale3'', ``scale2'', and ``scale1''. 
As shown in Fig. \ref{fig:attention}, with a growing number of EConv layers, the extracted features facilitate progressively more accurate dictionary queries.
\begin{figure*}[h]
    \label{fig:attention}
    \centering
    \includegraphics[width=0.95\textwidth]{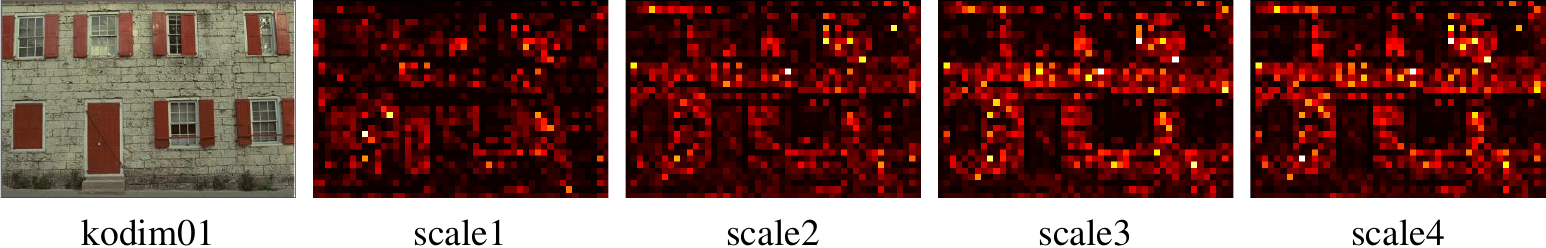}
    \caption{ Multi-Scale visualizations}
    \label{fig:attention}
\end{figure*}

\section{More visual examples}
\label{sec:More visual examples}
\begin{figure*}[]
    \centering
    \includegraphics[width=0.90\textwidth]{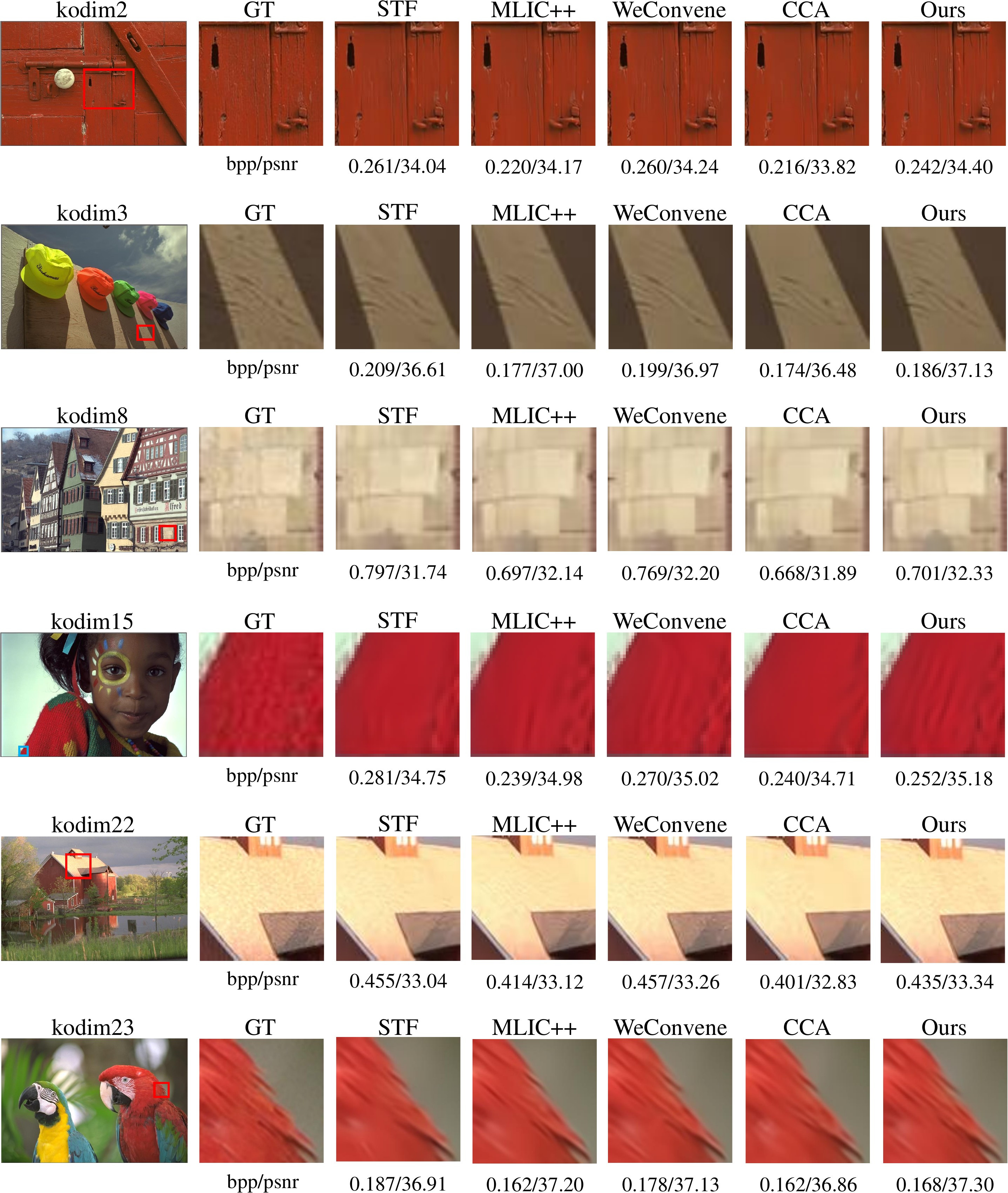}
    \caption{Reconstructed images from the Kodak dataset.}
    \label{fig:example_b}
\end{figure*}

Fig. \ref{fig:example_a} and Fig. \ref{fig:example_b} shows the  reconstruction results between our method and State-of-the-Art Methods.
Our model achieves better texture detail restoration at similar bpp levels.
For example, in Fig. \ref{fig:example_a}, our model more effectively restores the striped textures on the sails of the boat (kodim09) and the stitching details on the dress (kodim18).

\section{More rate-distortion results}
\label{sec:More rate-distortion results}
\begin{figure*}[]
    \centering
    \includegraphics[width=0.90\textwidth]{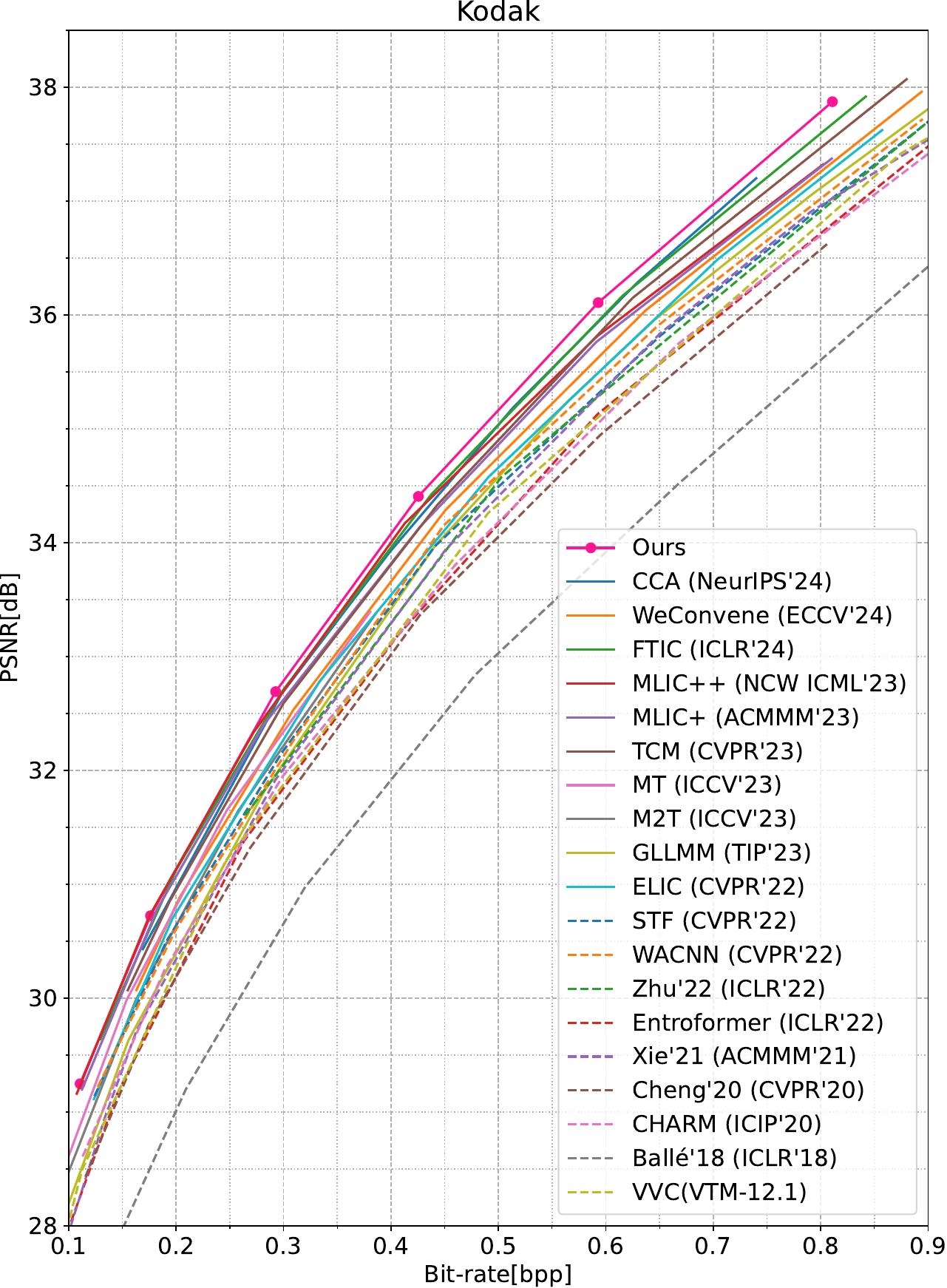}
    \caption{Performance evaluation (PSNR) on the Kodak dataset.}
    \label{kodak_psnr_all}
\end{figure*}

\begin{figure*}[]
    \centering
    \includegraphics[width=0.90\textwidth]{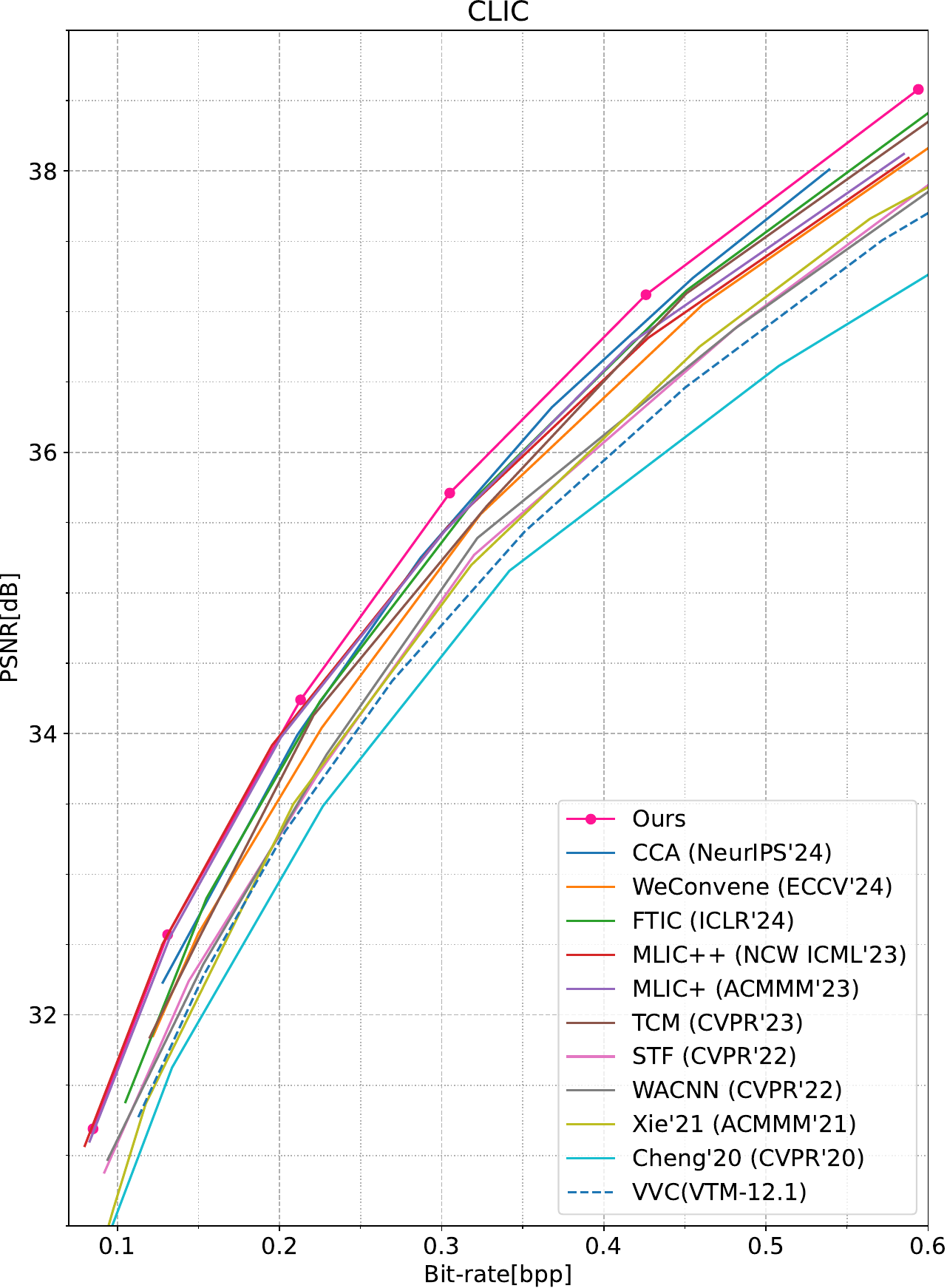}
    \caption{Performance evaluation (PSNR) on the CLIC dataset.}
    \label{clic_psnr_all}
\end{figure*}

\begin{figure*}[]
    \centering
    \includegraphics[width=0.90\textwidth]{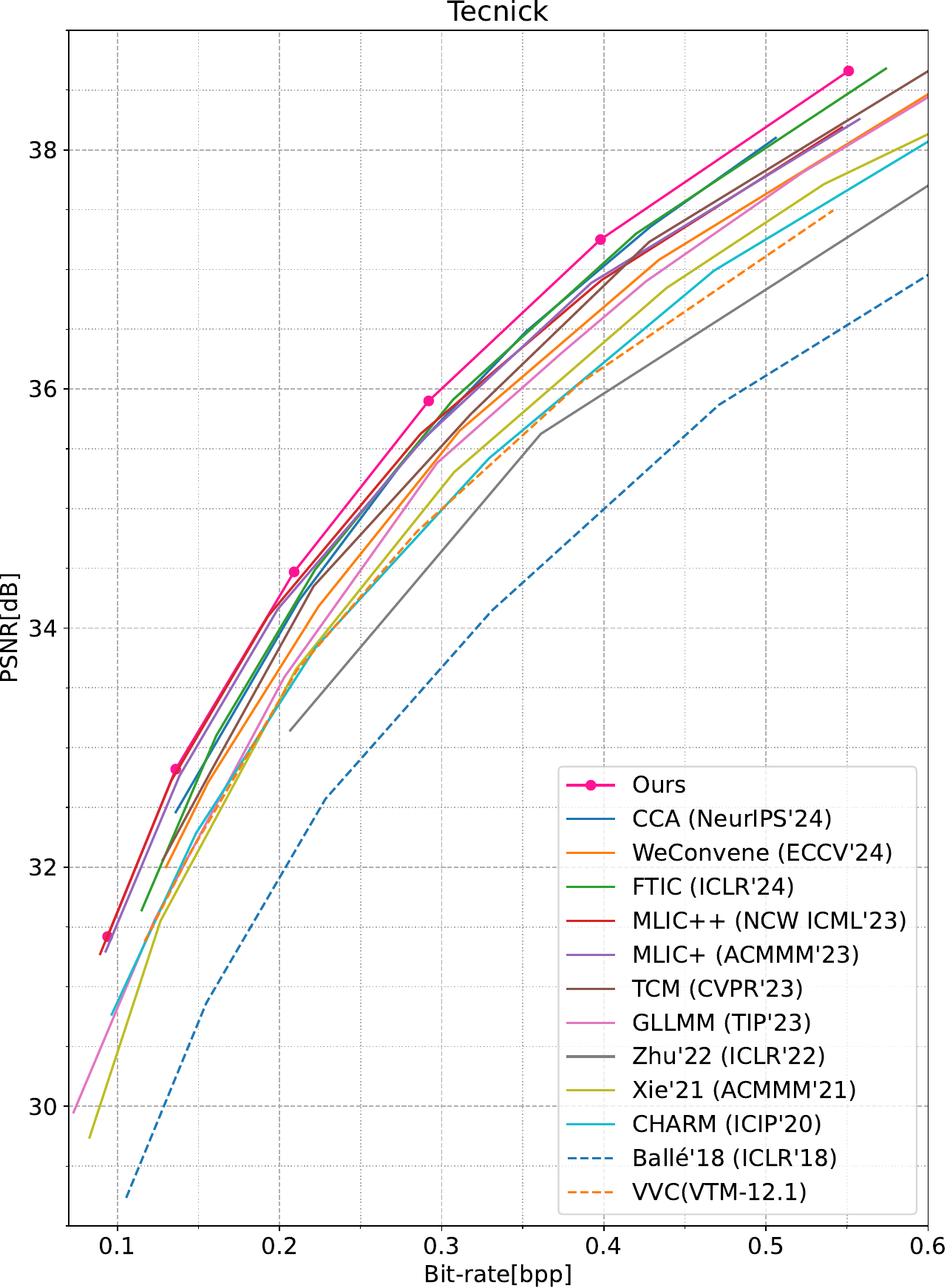}
    \caption{Performance evaluation (PSNR) on the Tecnick dataset.}
    \label{tecnick_psnr_all}
\end{figure*}

\begin{figure*}[]
    \centering
    \includegraphics[width=0.90\textwidth]{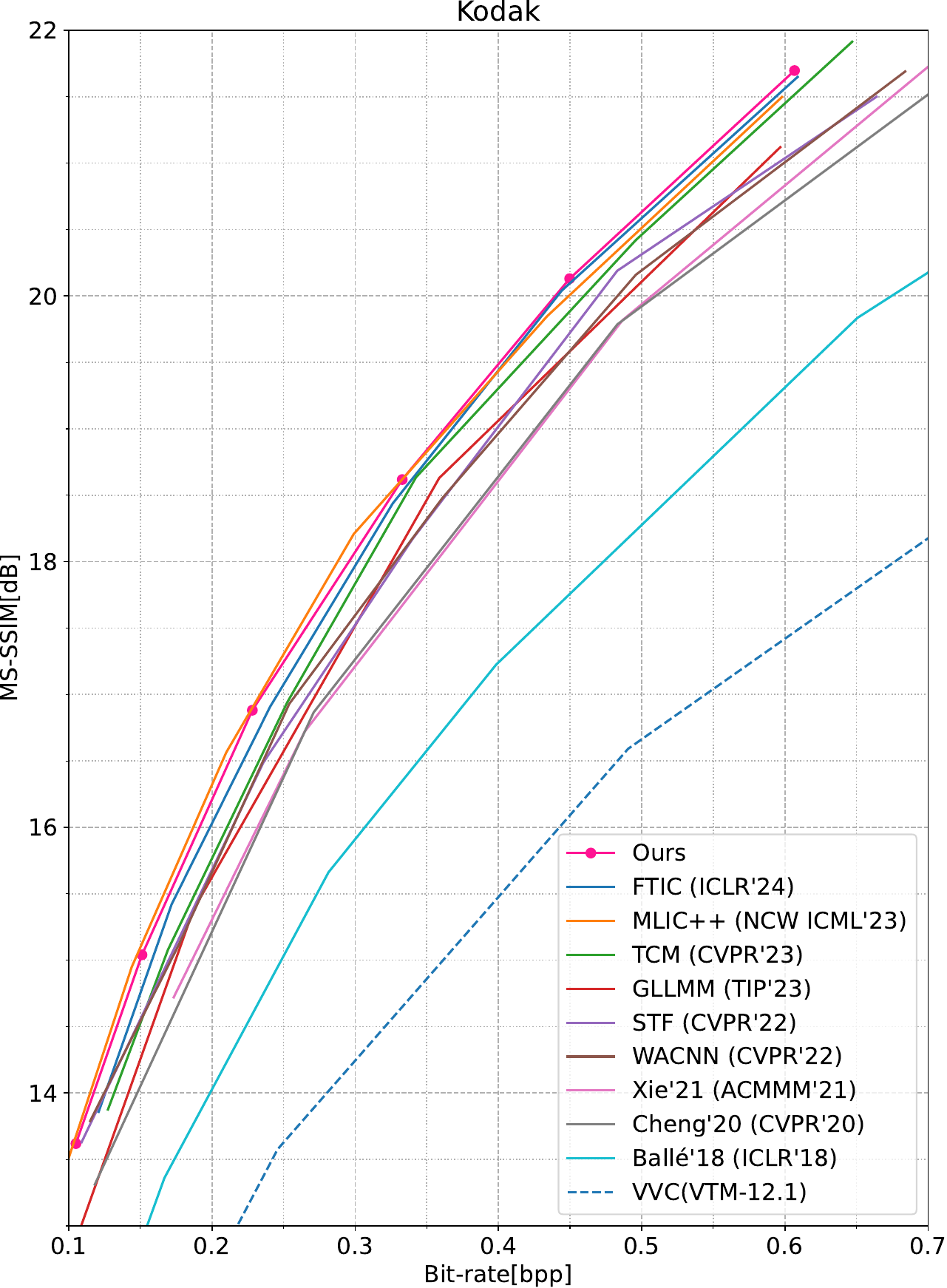}
    \caption{Performance evaluation (MS-SSIM) on the Kodak dataset.}
    \label{kodak_ssim_all}
\end{figure*}

For completeness, we present additional methods for comparison (Fig. \ref{kodak_psnr_all}, Fig. \ref{clic_psnr_all}, and Fig. \ref{tecnick_psnr_all}).
In addition, we also provide MS-SSIM optimized models (Fig. \ref{kodak_ssim_all}) to compare with other methods.
Our model achieves  state-of-the-art results on all datasets.

\begin{figure*}[ht]
    \centering
    \includegraphics[width=0.9\textwidth]{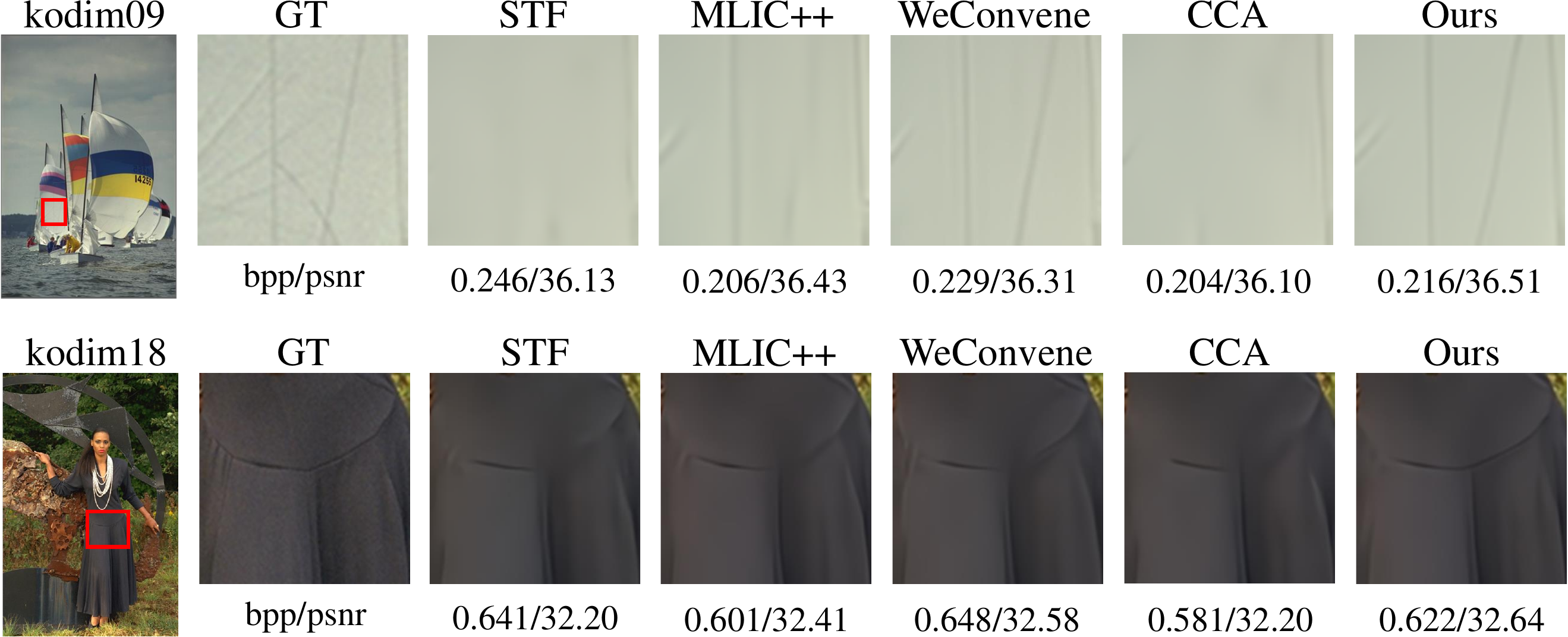}
    \caption{Reconstructed images from the Kodak dataset.}
    \label{fig:example_a}
\end{figure*}

